\documentclass{article}

\usepackage{amsmath}
\usepackage{capt-of}

\usepackage{PRIMEarxiv}
\usepackage[utf8]{inputenc} % allow utf-8 input
\usepackage[T1]{fontenc}    % use 8-bit T1 fonts
\usepackage{hyperref}       % hyperlinks
\usepackage{url}            % simple URL typesetting
\usepackage{booktabs}       % professional-quality tables
\usepackage{amsfonts}       % blackboard math symbols
\usepackage{nicefrac}       % compact symbols for 1/2, etc.
\usepackage{microtype}      % microtypography
\usepackage{fancyhdr}       % header
\usepackage{graphicx}       % graphics
\title{Style Transfer Dataset: What Makes A Good Stylization?
}

\author{
	Victor Kitov \\
	Lomonosov Moscow State University \\
	Plekhanov Russian University of Economics \\
	Moscow, Russia \\
	\texttt{v.v.kitov@yandex.ru} \\
\And
	Valentin Abramov \\
	Lomonosov Moscow State University \\
	Moscow, Russia \\
	\texttt{ab.val26@yandex.ru} \\
\And	
	Mikhail Akhtyrchenko \\
	Lomonosov Moscow State University \\
	Moscow, Russia \\
	\texttt{aht.misha@gmail.com} \\
}

\begin{document}
	\maketitle

	\begin{abstract}
We present a new dataset with the goal of advancing image style transfer - the task of rendering one image in the style of another image. The dataset covers various content and style images of different size and contains 10.000 stylizations manually rated by three annotators in 1-10 scale. Based on obtained ratings, we find which factors are mostly responsible for favourable and poor user evaluations and show quantitative measures having statistically significant impact on user grades. A methodology for creating style transfer datasets is discussed. Presented dataset can be used in automating multiple tasks, related to style transfer configuration and evaluation.
	\end{abstract}

\section{Introduction}

\begin{figure*}[h!t]
	\centering
	\includegraphics[width=1.0\textwidth]{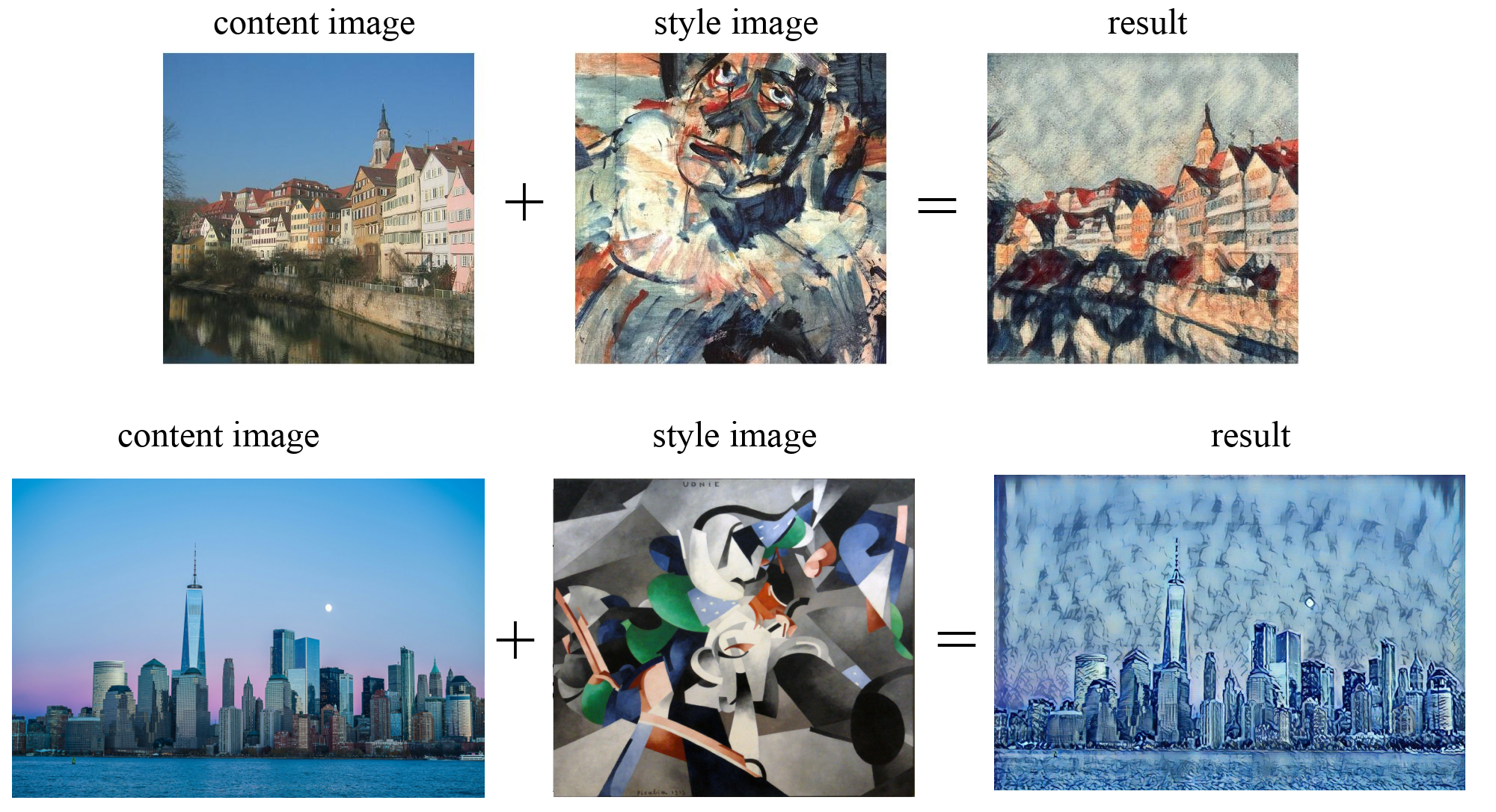}
	\caption{Style transfer task: redraw content image in the style of the style image transferring style colors as well (first row) and preserving original content colors (second row).}
	\label{fig:st_example}
\end{figure*}

Style transfer is an exciting research area where given a \emph{content image} (or content, e.g. a family photo) and a \emph{style image} (or style, e.g. a painting of a famous artist) the task is to redraw the content in the style of the style image, resulting in target stylization, as shown on Fig.~\ref{fig:st_example}. 

Style transfer automates artwork creation, drastically simplifying digital art creation, and can be applied in design, fashion, advertising industries, can facilitate social interaction by sharing vivid and memorable images. Many mobile apps  and online platforms provide style transfer service such as \hyperlink{https://prisma-ai.com}{prisma-ai.com} \hyperlink{https://www.ostagram.me}{ostagram.me} and \hyperlink{https://picsart.com/}{picsart.com}.

But what makes up a good and bad stylization? Since the final result is a digital artwork, it is hard to come up with a formal quality score and most studies rely on human evaluations when comparing different stylization methods. In such human assessments users are asked to rate stylizations on absolute scale or to select the best stylizations from a number of variants. Image evaluation program is usually re-implemented from scratch for such studies. The set of content and style images is specific for each user study which makes the experiments carried out non-reproducible and incomparable across articles.

To solve mentioned problems we propose a large style transfer dataset\footnote{\url{https://github.com/victorkitov/style-transfer-dataset}} with permissive license, containing contents, styles and manually evaluated stylizations.

To support future human evaluations of various image generation algorithms, we share our image evaluation program \footnote{\url{https://github.com/EnriFermi/image-evaluation-app}}. 

By training a stylization quality evaluation model on our dataset, researchers in style transfer would be able to automate stylization evaluations and make their results comparable across different articles.

In general, our dataset can be used for training models with various objectives:

\begin{itemize}
	\item given stylization, predict its user rating;
	\item given content, style and style size, predict how well they match each other;
	\item given content and style, predict optimal style size;
	\item given content and a sequence of styles, rank styles together with recommended style sizes from most matching to least matching.
\end{itemize}

Such tasks are useful in practical style transfer applications, automating best style image and size selection for user content.

Content images, forming the dataset, were carefully selected to cover a broad set of images, containing a diverse set of objects shown from different positions and under different lighting conditions. Styles were taken from github repositories of major style transfer methods. Stylizations were rated on absolute scale from 1 (lowest quality) to 10 (highest quality). Since artwork evaluation is subjective, each image was rated by three annotators, thus our dataset contains 30.000 grades in total. 

Common style transfer methods map not only artistic patterns, such as brushstrokes, but also color distribution from the style image. Examples of style transfer including and excluding color information are shown on Fig.~\ref{fig:st_example} above and below respectively. 

We argue that style transfer should map artistic patterns only, since transferring colors can be achieved by separate well-established methods, such as mean\&variance matching or histogram matching~\cite{gonzalez2009digital}. So we evaluated the clean style transfer mapping artistic patterns only without color information to omit the bias, favoring particular style colors.

\begin{figure*}[h!t]
	\centering
	\includegraphics[width=1.0\textwidth]{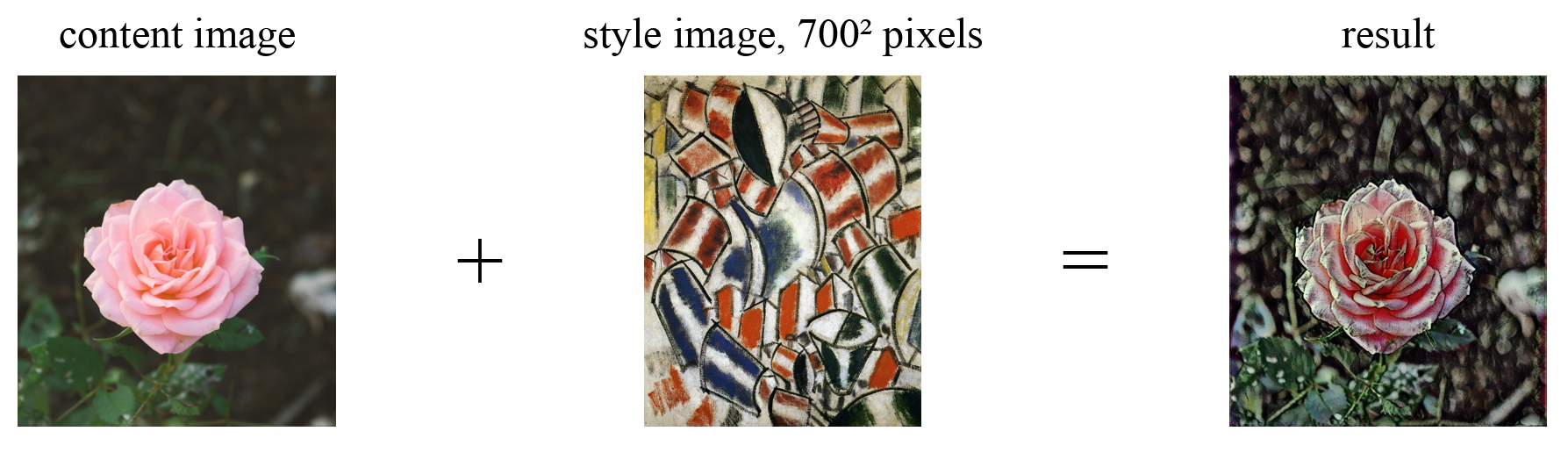}
	\includegraphics[width=1.0\textwidth]{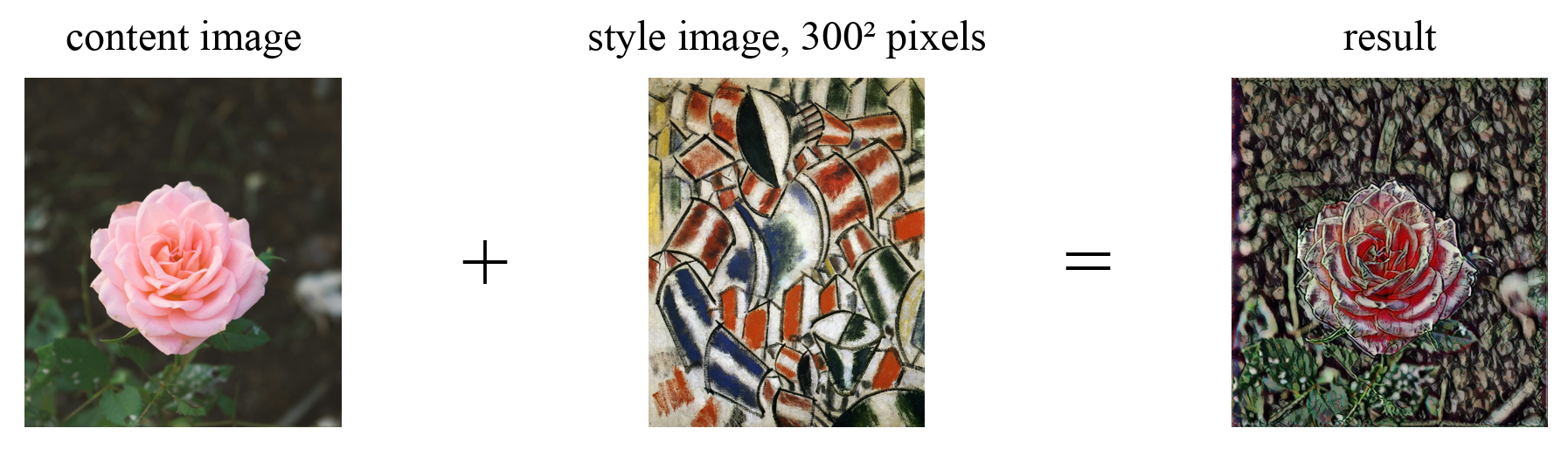}
	\caption{Style transfer using style image with different size: $700^2$ pixels (upper row) and $300^2$ pixels (lower row).}
	\label{fig:st_scale}
\end{figure*}

Also the size of the style image is an important factor, since all stylization methods are based on convolutions with fixed kernels. It was mentioned in~\cite{gatys2017controlling} and shown on Fig.~\ref{fig:st_scale}. Larger style size imposes larger brushstrokes and other artistic effects, so various style sizes were considered. 

Stylizations were performed by popular method ArtFlow~\cite{an2021artflow} providing similar results with other recent methods compared in~\cite{deng2022stytr2}, performing arbitrary style transfer by passing content and style image through the neural network. Similarity of style transfer methods can be explained by the fact, that although they utilize different architectures, their parameters are trained on the same loss function, proposed in~\cite{gatys2016image} or its slight variations, such as bringing closer centered covariance matrices between features in inner representations instead of uncentered ones or bringing closer only means and standard deviations instead of full covariance matrices - all these modifications target the same goal of bringing feature distributions of stylization image closer to feature distribution of style image, thus produce similar results. 

Applicability of our dataset is somewhat more limited to earlier style transfer methods, based on different principles and using different loss functions, namely
\begin{itemize}
	\item optimization-based style transfer~\cite{gatys2016image}, where optimization is performed not in the space of neural network weights, but in the space of pixels of the resulting stylization;
	\item patch-based style transfer~\cite{li2016combining},\cite{chen2016fast}, where stylization is performed by replacing content patches by most similar style patches.
\end{itemize}

We share our recipes for making a diverse and informative dataset for style transfer and encourage scientific community to extend the existing one.

Given a diverse set of rated stylizations, we also share our observations about style transfer performance and its limitations in general as well as recipes for making a high quality stylization. 

We also list common pitfalls of style transfer and find image features, having statistically significant impact on artwork perception and evaluation by the users. We hope that our findings will highlight directions for future improvements in style transfer methods.

Overall the contribution of our work is the following:
\begin{itemize}
	\item we provide a public dataset with permissive license that can be used for creating an overall picture how style transfer works in different contexts and can be used for building models, assisting in automatic stylization evaluation and optimal style and style size selection;
	\item we discuss our methodology for creating diverse and informative style transfer dataset;
	\item we share general observations under what conditions style transfer methods work well or produce poor results.
	\item quantitative image measures are found, having statistically significant impact on final user evaluation.
\end{itemize}

The paper is structured as follows. Section~\ref{seq:related_work} describes related work.
Section~\ref{seq:data_collection} describes data collection and preparation process.
Section~\ref{seq:analysis} provides general analysis of obtained ratings. In section~\ref{seq:impact-factors} we give our insights which qualitative factors contribute to good and bad stylization and describe quantitative factors, having statistically significant impact on user ratings. In section \ref{seq:recommendations} we provide a shortlist of recommendations for generating a high-quality stylizations and section~\ref{seq:conclusion} concludes.
 
\section{Related Work} \label{seq:related_work}
To the best of our knowledge, there was only one attempt to create a large public style transfer dataset in~\cite{chen2022quality}. Although we acknowledge the importance of the work conducted in~\cite{chen2022quality}, in our view it has some limitations, which we tried to overcome in our research.
\begin{itemize}
	\item Only 150 content-style image pairs were evaluated, which is insufficient to train even simple neural networks. Our dataset contains evaluations for 10.000 content-style pairs.
	\item Content images were rescaled and cropped to $512 \times 512$ resolution. We argue that the end user is interested in stylizing content images of higher resolution without the restriction, that this image should be square. Thus content images in our dataset are HD and preserve original aspect ratio.
	\item Style images were also set to $512 \times 512$ resolution. Although this is a common strategy to rescale style images to this shape in scientific articles, the end user is interested to apply style transfer to styles of arbitrary size and shape. Besides we stress that style size has a major impact on the quality of the resulting stylization as shown on Fig.\ref{fig:st_scale}. Thus we do not put any constraints on aspect ratio and consider different sizes of each style image, containing $150^2,300^2,500^2$ and $700^2$ pixels. We show that the size of style image affects the size of imposed style patterns and has a major effect on stylization quality.
	\item Content-style pairs in~\cite{chen2022quality} were carefully selected to match each other. We argue, that the end users might want to mix any pair of content and style and frequently this produces favorable results, even if original content and style images do not match each other on semantic level, as can be seen, for example, on stylizations from \hyperlink{https://www.ostagram.me}{ostagram.me} website.
	\item Style transfer methods, that were evaluated in~\cite{chen2022quality}, were proposed in 2017-2020. We consider only ArtFlow~\cite{an2021artflow} method, proposed in 2021 and it produces qualitatively similar results to later methods as can be seen in~\cite{deng2022stytr2} and our general observations beyond the scope of this study.
	\item In~\cite{chen2022quality} annotators were asked to select a better stylization out of two alternatives. While we agree, that such approach produces less bias, we asked each annotator to rate each stylization individually in 1-10 scale, because such approach extracts much more information from each annotation. In $N$ pairwise evaluations  only $N$ pairwise comparisons could be extracted, while $N$ individual evaluations of each image produce $N^2$  pairwise comparisons, since every image can be compared to every other on absolute scale.
\end{itemize}

\section{Dataset Creation Process} \label{seq:data_collection}
Next we describe how proposed dataset was created.

\subsection{Content and Style Images}

The dataset was made using 50 contents and 50 style images. Since style size strongly effects stylization result, as shown on Fig.~\ref{fig:st_scale}, style images were rescaled (keeping aspect ratio) to contain $150^2, 300^2, 500^2$ and $700^2$ pixels. We did not use any preliminary matching between content and style images, because frequently even semantically different contents and styles produce expressive results. So instead we stylized every combination of content, style and style size, obtaining $50\cdot 50 \cdot 4=10.000$ stylizations. At the end some combinations work well together, some not and our dataset allows to evaluate (and train models of automatic evaluation) how well content and style match each other and which style size is the best.

Since artistic tastes of different people differ, each stylization was rated by three annotators, thus containing $30.000$ grades. This enables to judge the level of disagreement between different annotators.

Content images cover diverse photos with people, animals, buildings and other common objects such as trains, cars, airplanes, trees and flowers. Images were selected to cover various lighting conditions (day, evening, night) and various distances to objects - close, medium and far distance. Landscapes and street views were also included.

Most content images were downloaded from \mbox{\url{unsplash.com}} with permissive license\footnote{\url{https://unsplash.com/license}}. Since unsucessful style transfer degrades the quality of photographs, portrait photographs of people were generated using a generative adversarial network~\cite{karras2017progressive} (contents 14 and 17) and a diffusion model~\cite{rombach2021highresolution} (contents 8, 13, 22, 26, 48) and upscaled\footnote{Upscaling with \url{https://www.pixelcut.ai/image-upscaler}} without losing photorealism. All content images were rescaled, keeping aspect ratio, to contain $900^2$ pixels following the idea, that final applications of style transfer are made in HD resolution.

Style images were taken from github repositories of popular style transfer methods, mainly ArtFlow\footnote{\url{https://github.com/pkuanjie/ArtFlow/tree/main/data/style}}, since these images are massively reused in style transfer community, produce good results and can be freely used for research purposes. A general observation is that style transfer methods work well with style images containing large brush strokes and other vivid artistic patterns.

\subsection{Stylization Creation}
After obtaining 50 content images, 50 style images of 4 different scales, 10.000 stylizations were generated using ArtFlow~\cite{an2021artflow} algorithm using official implementation\footnote{\url{https://github.com/pkuanjie/ArtFlow}}, which is relatively new, well-recognized in research community. It produces high-quality results on average, works with contents and styles of different sizes with arbitrary aspect ratio and is economical in terms of computing resources (can be trained and is able to produce HD results on GPU even with 11GB memory. We used an official pretrained model.

Implementations of more recent style transfer algorithms, namely AdaAttN~\cite{liu2021adaattn}\footnote{\url{https://github.com/Huage001/AdaAttN}} and Stytr2~\cite{deng2022stytr2}\footnote{\url{https://github.com/diyiiyiii/StyTR-2}} are also able to produce high quality results, however 
\begin{enumerate}
	\item require much more memory (11GB memory in GPU is not enough to produce HD results),
	\item official implementations work only when content and style have equal size (user may want to vary image sizes),
	\item require content and style image to have square size (actual image sizes are rectangular).
\end{enumerate}
Because of these limitations, we stick to ArtFlow method on our analysis. Besides, the comparative analysis of stylizations obtained by ArtFlow, AdaAttN and Stytr2 show qualitatively similar results in visualizations~\cite{deng2022stytr2}.
We did not consider generative adversarial nets and diffusion models, capable of performing style transfer, since they are much more demanding in terms of computational resources and we were interested in the specific field of style transfer models, which are easy to train and lightweight.

\subsection{Style Recoloring}
Different styles have different color distributons. Standard style transfer applies not only drawing patterns (like brush strokes), but color distribution as well. This may lead to inconsistent results, e.g. redrawing a bright photo with dark colors. It can also lead to color bias in user evaluations (some colors are more preferable by respondents. To omit both problems we transfer only style information without color information. To achieve that style image was recolored to the colors of the content image before stylization, using mean\&variance matching in LAB color space, namely using the following algorithm:
\begin{enumerate}
	\item Convert content and style image from RGB to LAB color scheme.
	\item Calculate means $\mu^s_l,\mu^s_a,\mu^s_b$ for LAB color channels of the style image.
	\item Calculate means $\mu^c_l,\mu^c_a,\mu^c_b$ for LAB color channels of the content image.
	\item Calculate standard deviations $\sigma^s_l,\sigma^s_a,\sigma^s_b$ for LAB color channels of the style image.
	\item Calculate standard deviations $\sigma^c_l,\sigma^c_a,\sigma^c_b$ for LAB color channels of the content image.
	\item Rescale LAB color channels $l^s,a^s,b^s$ of the style image to bring color distribution closer to the content image:
	\begin{align*}
		l^s &:= \frac{\sigma^c_l}{\sigma^s_l} (l^s-\mu^s_l)+\mu^c_l \\
		a^s &:= \frac{\sigma^c_a}{\sigma^s_a} (a^s-\mu^s_a)+\mu^c_a \\
		b^s &:= \frac{\sigma^c_b}{\sigma^s_b} (b^s-\mu^s_b)+\mu^c_b \\
	\end{align*}	
\end{enumerate}

\begin{figure*}[h!t]
	\centering
	\includegraphics[width=1\textwidth]{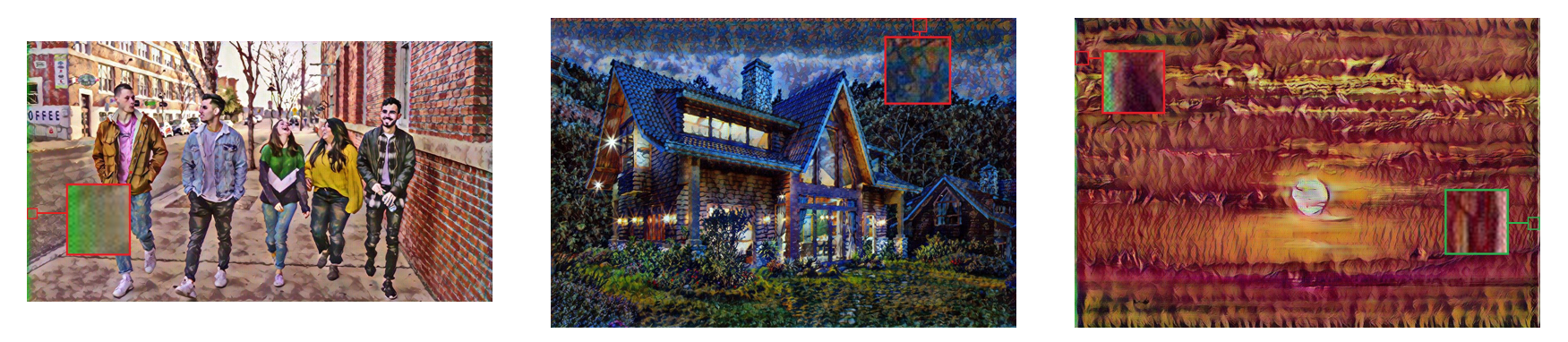}
	\caption{Common border artifact present in stylizations by ArtFlow style transfer method.}
	\label{fig:border_artifact}
\end{figure*}

\begin{figure*}[h!t]
\begin{minipage}{.45\textwidth}
  \centering
  \includegraphics[width=1\linewidth]{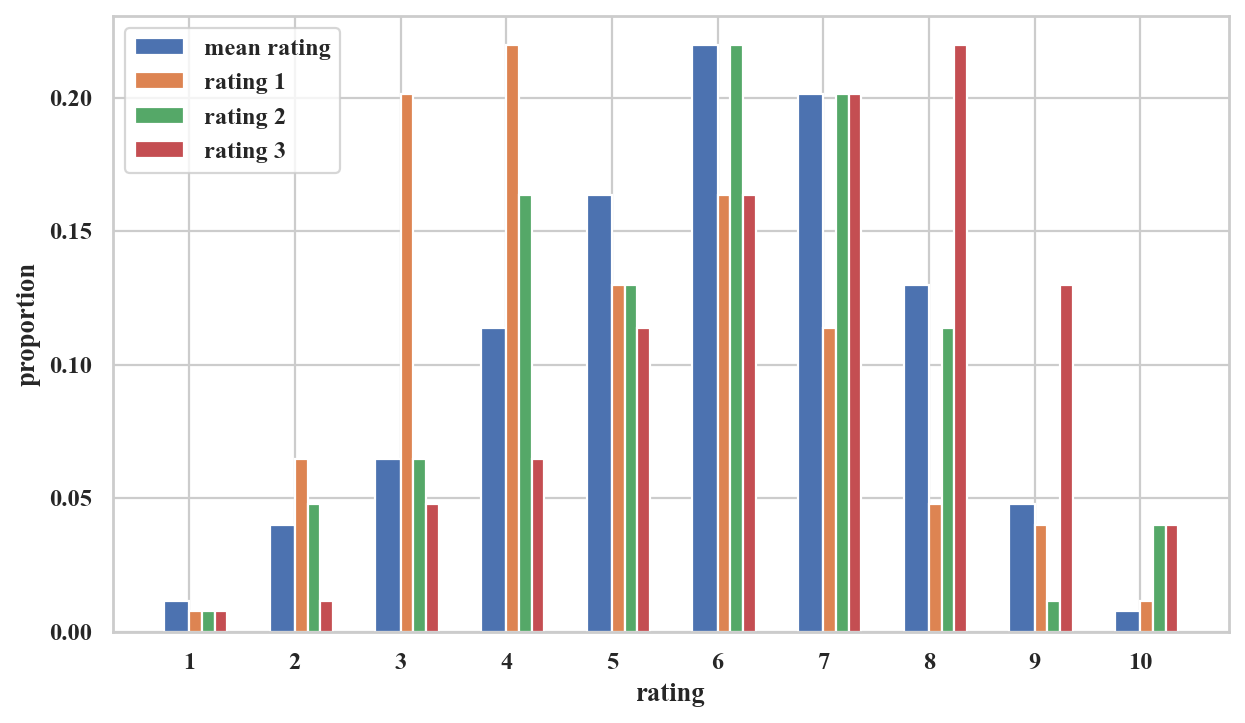}
  \captionof{figure}{Rating distribution for 3 annotators and mean rating.}
  \label{fig:rating_distribution}
\end{minipage}%
\hspace{1cm}
\begin{minipage}{.45\textwidth}
  \centering
  \includegraphics[width=1\linewidth]{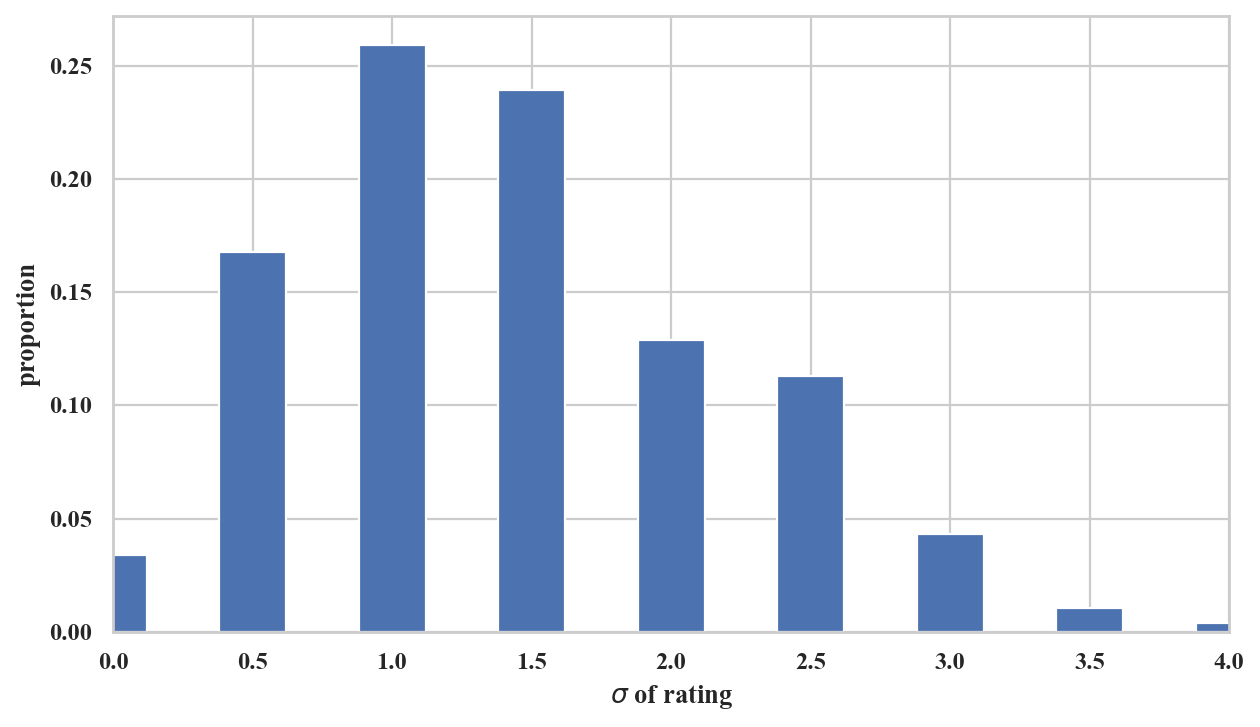}
  \captionof{figure}{Distribution of standard deviations of ratings. }
  \label{fig:rating_std_distribution}
\end{minipage}
\end{figure*}

Recoloring was performed in LAB color space, since this color space is more natural: uniform changes of components in the LAB color correspond to uniform changes in perceived color by human eye\footnote{\url{https://en.wikipedia.org/wiki/CIELAB_color_space}}. 
We also experimented with histogram matching \footnote{\url{https://en.wikipedia.org/wiki/Histogram_matching}}, using \texttt{skimage} package, but it performed worse on content and style image pairs. 

\subsection{Labeling Process}

Each of 10.000 stylizations was labeled by three annotators on scale from 1 (worst) to 10 (best), giving 30.000 ratings, using a software program, specifically designed for human image evaluation. 

During labeling process, stylizations were shown to each annotator in full screen. Annotators were asked to cover the whole range of grades from 1 to 10 dateset-wide, however, there were no restrictions to cover this range for particular content and style. Indeed, some contents and styles yielded good and some --- bad results on average.

Only stylizations were shown for evaluation, as the end user of style transfer systems cares mostly about final result. Annotators were instructed to grade them according to their subjective aesthetic pleasure, by "willingness to use it as a picture on the wall or as an illustration on a website of particular topic". Annotators were asked not to take the actual content intro account, to account only for artistic expressiveness of the picture and its personal appeal. For particular styles stylization produced photorealistic result without any artistic effect. In such cases annotators were asked to grade stylization in the range 4-6, depending on color vividness and presence of local style transfer artifacts.

Each annotator first rated all stylizations of one content image, than of another one, etc. This was made to make it easier for annotators to compare relative stylization quality of the same content image. After each evaluation, a large grade was shown on top of evaluated image for 0.5 sec. to help annotators remember how they evaluated each kind of stylization. The order of contents was random among annotators to compensate for hypothetical trends in grades as the evaluation process progressed.

It was noted, that ArtFlow frequently produces artifacts at the borders of the stylization image, as shown on Fig.~\ref{fig:border_artifact}.

Annotators were asked not to downgrade stylizations for the presence of this particular artifact, since it can be easily removed by slight border cropping. All other artifacts were downgraded.

\begin{table*}[h!t]
	% \centering
	\begin{minipage}{1.0\linewidth}
		\centering
		\begin{tabular}{|l|l|l|l|}
			\hline
			& rating 1 & rating 2 & rating 3 \\ \hline
			rating 1 & 1.0000   & 0.4030   & 0.3612   \\ \hline
			rating 2 &          & 1.0000   & 0.4314   \\ \hline
			rating 3 &          &          & 1.0000   \\ \hline
		\end{tabular}
		\caption{Kendall's Tau-B correlations between ratings, showing statistically significant partial monotone relationship between ratings of 3 annotators.}
		\label{table:kendall_corr}
	\end{minipage}
\end{table*}

\begin{table*}[h!t]	
	\begin{minipage}{1.0\linewidth}
		\centering
		\begin{tabular}{|l|l|l|l|}
			\hline
			measure         & $\sigma(L)$     & $\sqrt{\sigma(L)^2 + \sigma(A)^2 + \sigma(B)^2}$     & sharpness score      \\ \hline
			Kendall's Tau-B & 0.0846 & 0.0941 & -0.2258 \\ \hline
			p-value         & 3e-35  & 3e-43  & 2e-253  \\ \hline
		\end{tabular}
		\caption{Kendall's Tau-B correlations between mean rating and different measures.}
		\label{table:kendall_corr2}
	\end{minipage} 
\end{table*}

\begin{figure*}[h!t]
	\centering
	\includegraphics[width=1\textwidth]{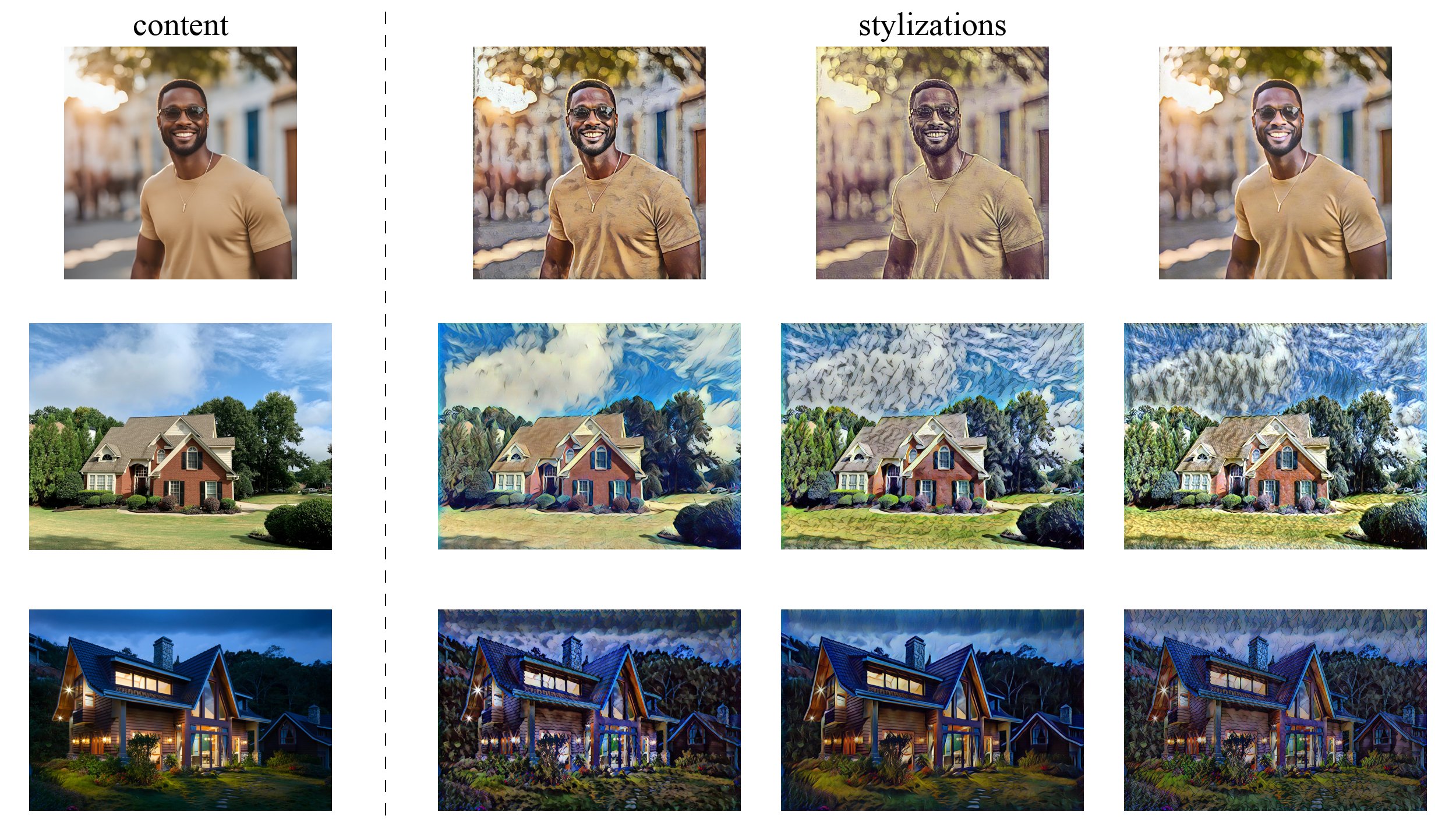}
	\caption{Examples of content images, yielding high quality stylizations.}
	\label{fig:content_good_stylizations}
\end{figure*}
 
\begin{table*}[h!t]
\centering
\begin{tabular}{|l|lllll||lllll|}
\hline
            & \multicolumn{5}{l||}{worst}                                                                                           & \multicolumn{5}{l|}{best}                                                                                            \\ \hline
content №   & \multicolumn{1}{l|}{14}   & \multicolumn{1}{l|}{17}   & \multicolumn{1}{l|}{36}   & \multicolumn{1}{l|}{43}   & 19   & \multicolumn{1}{l|}{4}    & \multicolumn{1}{l|}{20}   & \multicolumn{1}{l|}{5}    & \multicolumn{1}{l|}{3}    & 26   \\ \hline
mean rating & \multicolumn{1}{l|}{4.6}  & \multicolumn{1}{l|}{4.64} & \multicolumn{1}{l|}{4.9}  & \multicolumn{1}{l|}{4.91} & 5.06 & \multicolumn{1}{l|}{6.44} & \multicolumn{1}{l|}{6.46} & \multicolumn{1}{l|}{6.6}  & \multicolumn{1}{l|}{6.62} & 6.65 \\ \hline
std rating  & \multicolumn{1}{l|}{1.41} & \multicolumn{1}{l|}{1.5}  & \multicolumn{1}{l|}{1.48} & \multicolumn{1}{l|}{2.2}  & 1.51 & \multicolumn{1}{l|}{1.34} & \multicolumn{1}{l|}{1.26} & \multicolumn{1}{l|}{1.42} & \multicolumn{1}{l|}{1.22} & 1.07 \\ \hline
\end{tabular}
\caption{5 contents with the least rating and 5 contents with the greatest rating in the dataset.}
\label{table:content_ratings}
\end{table*}

\section{General Analysis of Ratings} \label{seq:analysis}

Distributions of individual ratings by each of three annotators, as well as distribution of mean grades (for each stylization) are shown on Fig.~\ref{fig:rating_distribution}. Mean rating has skewness towards higher ratings. Individual ratings have different distributions, which confirms the hypothesis that different people have different aesthetic preferences. Although the rating distributions differ, the standard deviation of the ratings mostly does not exceed two points, as seen on Fig.~\ref{fig:rating_std_distribution}. To analyze the correlations between ratings, we used Kendall rank correlation\footnote{\url{https://en.wikipedia.org/wiki/Kendall_rank_correlation_coefficient}}, evaluating the strength of monotone dependencies between ratings irrespective of the kind of those dependencies. More precisely we measured Kendall's Tau-B measure implementation from scipy library\footnote{\url{https://scipy.org/}}. To test for statistical significance, we performed statistical test with null hypothesis, that the rank correlation is zero with a two-sided alternative ~\cite{scipy} ~\cite{kendall-tau-b}, which was rejected with high confidence. The correlations in table ~\ref{table:kendall_corr} show that the ratings of each annotator have a largely monotonic dependence between each other, and thus we can use the averaged ratings to evaluate the general quality of each stylization.

In Table ~\ref{table:content_ratings} shows that the quality of the stylization highly depends on the content image: some content images have on average lower average rating than others.

\section{Impact Factors} \label{seq:impact-factors}
In this section we analyze some of the most important factors affecting user ratings.

\begin{figure*}[h!t]
	\centering
	\includegraphics[width=1\textwidth]{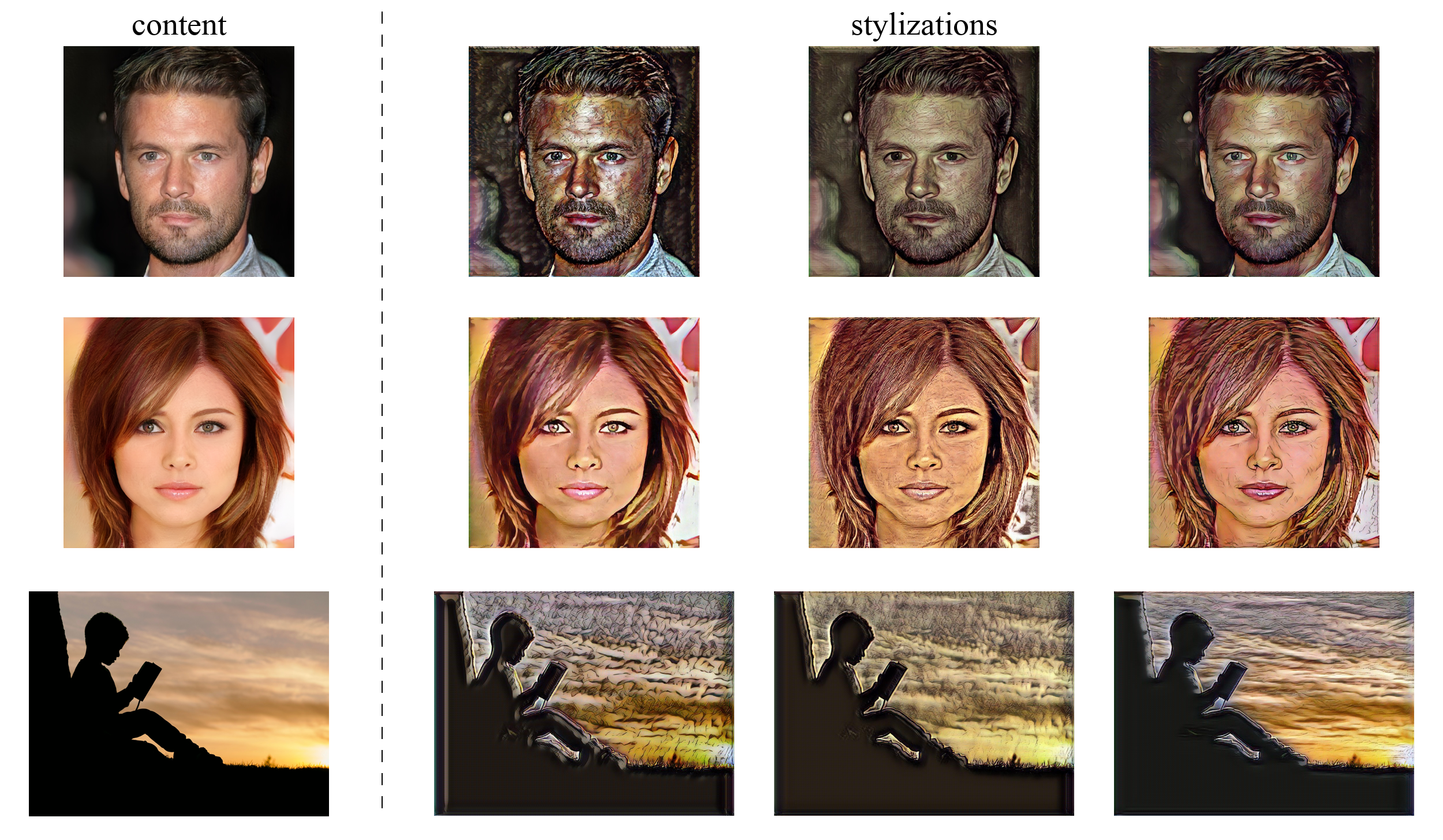}
	\caption{Examples of content images, yielding low quality stylizations.}
	\label{fig:content_bad_stylizations}
\end{figure*}
\begin{table*}[h!t]
% \centering
\centering
\end{table*}

\begin{figure*}[h!t]
	\centering
	\includegraphics[width=1\textwidth]{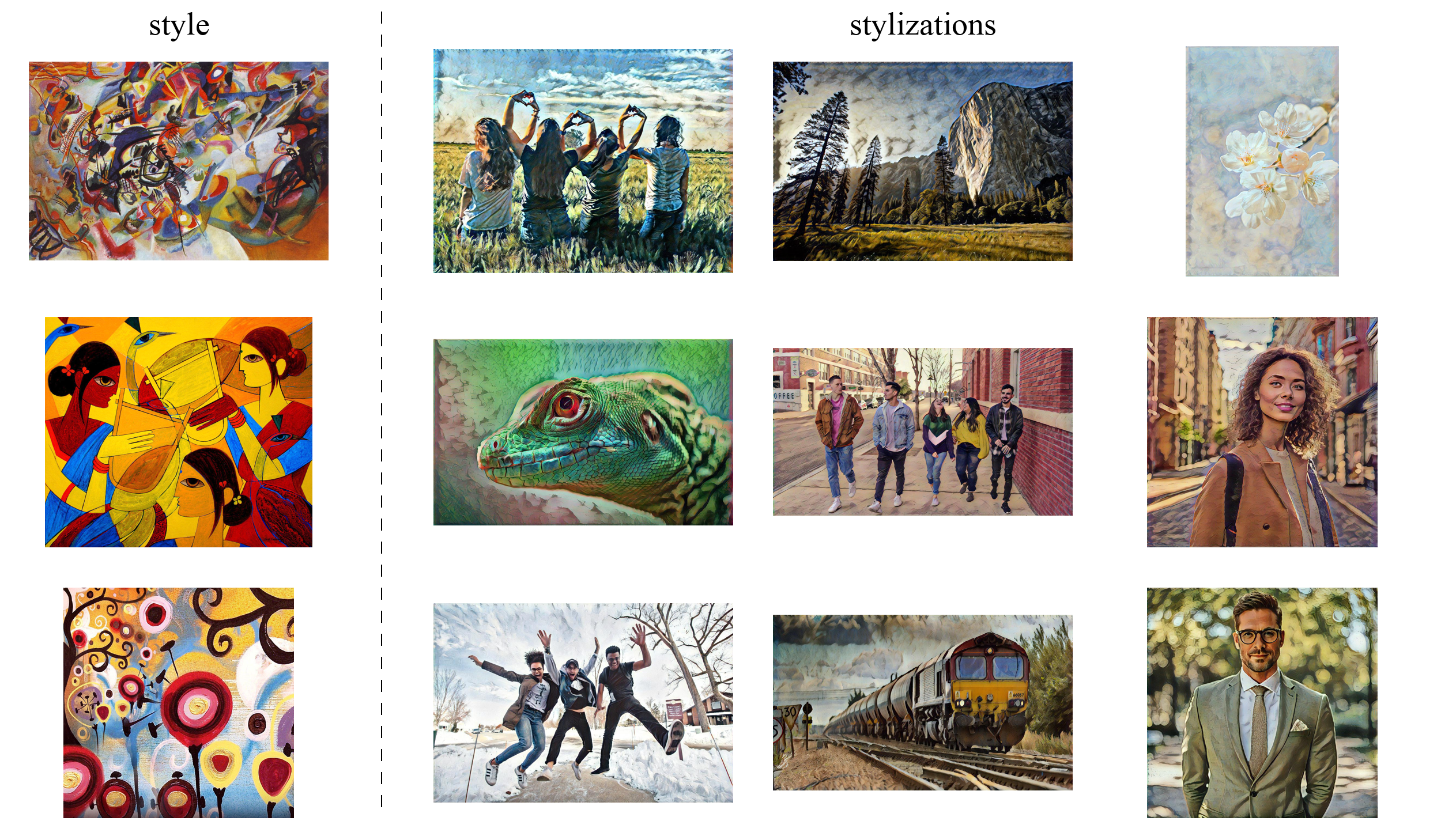}
	\caption{Examples of styles that produce high quality stylizations. Respective sizes of style images are $300^2$, $500^2$ and $700^2$ pixels.}
	\label{fig:style_good_stylizations}
\end{figure*}

\begin{table*}[h!t]
% \centering
\begin{minipage}{.45\linewidth}
\centering
\begin{tabular}{|l|l|l|l|l|l|l|l|}
\hline
measure         &  $\rho(I_{c}, I)$ \\ \hline
Kendall's Tau-B & -0.4696 \\ \hline
p-value         & 1e-259 \\ \hline
\end{tabular}
\caption{Kendall's Tau-B correlations between mean rating and $\rho(I_{c}, I)$ with independence null hypothesis. This analysis was restricted to pairs from dataset for which $\text{Faces}(I_{c}) \neq \emptyset$.}
\label{table:kendall_corr3}
\end{minipage}
\hspace{1cm}
\begin{minipage}{.45\linewidth}
\centering
\begin{tabular}{|l|l|l|l|l|}
\hline
style size (px)  & $150^2$ & $300^2$ & $500^2$ & $700^2$ \\ \hline
mean rating & 5.19 & 5.95 & 6.11 & 6.09 \\ \hline
std rating  & 1.43 & 1.46 & 1.51 & 1.55 \\ \hline
\end{tabular}
\caption{Style sizes and their respective rating statistics.}
\label{table:size_ratings}
\end{minipage} 
\end{table*}

\subsection{Color Diversity and Sharpness}

Highly rated stylizations have several common features: they do not heavily distort the shapes of the objects on the image, but add local textures from the style image to the background. The quality of stylizations also depends on the color and the brightness diversity of the resulting image. To compute the diversity, we convert the stylizations to the LAB color space and compute the brightness diversity as standard deviation of the luminance channel $\sigma(L)$ and color diversity as $\sqrt{\sigma(L)^2 + \sigma(A)^2 + \sigma(B)^2}$. The Kendall's Tau-B correlation coefficients in Table ~\ref{table:kendall_corr2} show that a correlation between mean ratings and these measures exists and is statistically significant (tested null hypothesis was absence of any dependency).

The mean sharpness of the stylization also affects the ratings. We compute the sharpness of the image as the variance of the image Laplacian (the result of applying the discrete Laplace filter to the image) as proposed in Pech-Pacheco et al.~\cite{laplacian-sharpness}. Correlation of this measure with ratings is negative, as shown in the Table ~\ref{table:kendall_corr2}: if the stylization is too sharp, the overall quality is lower. 

Overly sharp images were usually produced by using style images that contained too many small and sharp patterns. These patterns overload content image with stylistic details and make final stylization difficult to perceive, as demonstrated on Fig.~\ref{fig:style_bad_stylizations} (middle row).

\subsection{Texture Blur and Edges Reproduction}

\begin{figure*}[h!t]
	\centering
	\includegraphics[width=1.0\textwidth]{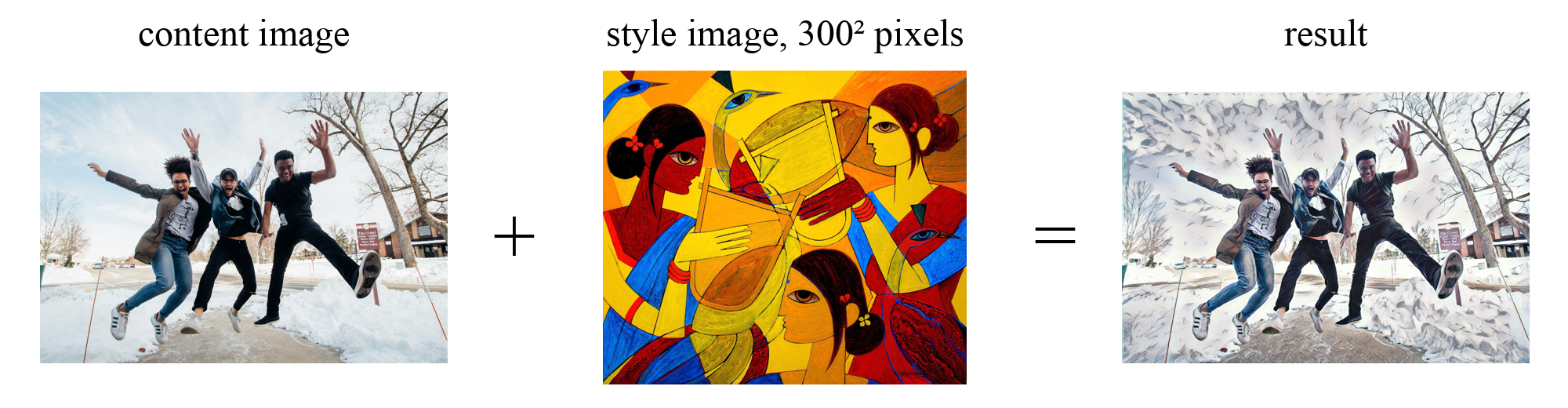}
	\includegraphics[width=1.0\textwidth]{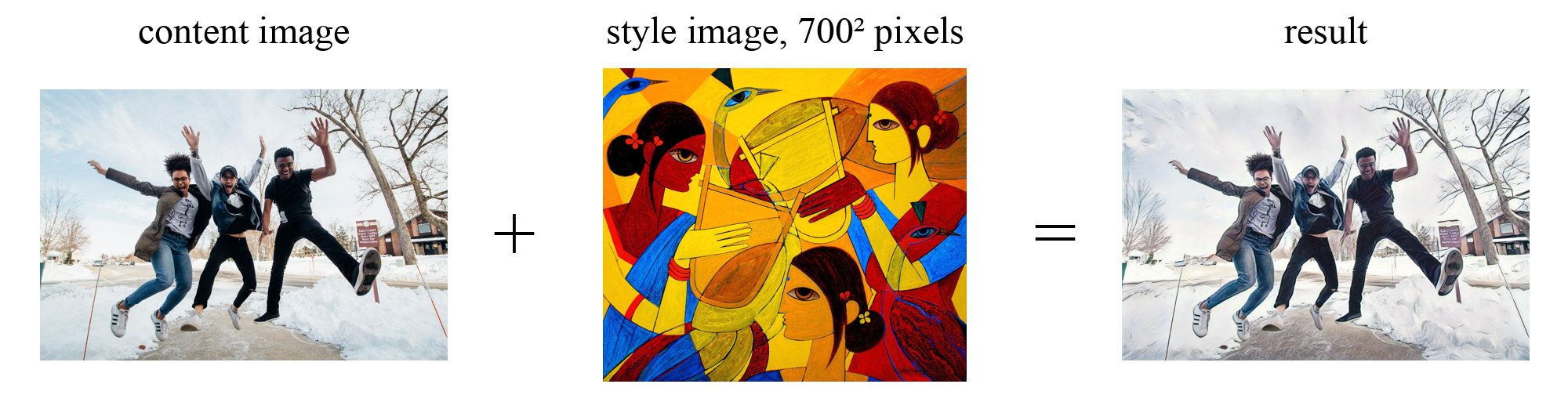}
	\caption{Style transfer fails to capture large artistic patterns for style images of larger size.}
	\label{fig:st_scale_2}
\end{figure*}

Stylizations with low ratings show that the style transfer process highly depends on the content images. The contents with the lowest average ratings are present on the Fig.~\ref{fig:content_bad_stylizations}. Full-face portraits are rated lower than other pictures. This happens because of distortion of the facial features caused by style transfer, especially on human skin --- even minimal wrinkles are amplified by style transfer as can be seen on two top rows of Fig.~\ref{fig:content_bad_stylizations}.

Another common feature in badly stylized contents is the presence of uniformly or smoothly colored regions as the black figure and the sky on the bottom row of~Fig.~\ref{fig:content_bad_stylizations}, which are frequently distorted by occasional style patterns, making them non-smooth.

\subsection{Face Reproduction}
As mentioned above, full human faces are generally very sensitive to style transfer. Small changes in facial features, such as eyes or nose lead to a decrease in the recognizability of a particular person's face in the image, which plays an important role in the case of subjective perception of stylization. We found a strong anti-correlation of the special distance function between faces on the content and stylized image with the stylization quality scores. The distance used was the cosine distance between the embeddings of the face areas of content and stylizations. The embeddings were obtained using the VGG-face model proposed by ~\cite{parkhi2015deepface} and implemented in the DeepFace library\footnote{\url{https://github.com/serengil/deepface}}.

Define content image with height $H_{c}$ and width $W_{c}$ as $I_{c}$ and define stylization image, having the same size, as $I$.
Let $X^{xyhw}$ denote rectangular fragment of image $X$, covered by rectangle with top-left corner $(x,y)$, width $w$ and height $h$. 

For each content image, we manually identified the set of faces present. For each detected face, we determined a rectangular bounding box defined by its coordinates and dimensions. 

Let $\text{Faces}(I_{c})$ denote the set of vectors $(x,y,h,w)$, where $(x,y)$ represents the coordinates of the top-left corner of the bounding box, and $(h,w)$ specifies its height and width. For each identified face in $I_{c}$ $\text{Faces}(I_{c})$ includes exactly one element describing bounding box of that identified face.

Define last layer output of VGG-Face model applied to some image $X$ as $\text{emb}(X) \in \mathbb{R}^{2622}$. We consider the following distance function: 

\begin{equation*}
\rho(I_{c}, I) = \\ \frac{1}{|\text{Faces}(I_{c})|}  \sum_{(x, y, h, z) \in \text{Faces}(I_{c})} \\ 1 - \frac{\langle \text{emb}(I_{c}^{xyhw}), \text{emb}(I_{s}^{xyhw}) \rangle}{\|  \text{emb}(I_{c}^{xyhw}) \|_2 \cdot  \| \text{emb}(I_{s}^{xyhw}) \|_2}
\end{equation*}
Next we evaluate Kendall correlation coefficient between defined distances and ratings for stylizations, containing human faces. We found that this correlation is strongly negative, and the null hypothesis of absence of correlation has negligible p-value, confirming, that original face reproduction is essential for high quality stylization. Exact values are shown in Table~\ref{table:kendall_corr3}.

\subsection{Style Size and Characteristics}
%\ref{table:size_ratings} - ratings dependence from style size
%\ref{table:style_ratings} - top and bottom 5 styles and sizes.

\begin{table*}[h!t]
	\centering
	\begin{tabular}{|l|lllll||lllll|}
		\hline
		& \multicolumn{5}{l||}{worst}                                                                                                          & \multicolumn{5}{l|}{best}                                                                                                           \\ \hline
		style №     & \multicolumn{1}{l|}{7}       & \multicolumn{1}{l|}{7}       & \multicolumn{1}{l|}{7}       & \multicolumn{1}{l|}{38}      & 7       & \multicolumn{1}{l|}{43}      & \multicolumn{1}{l|}{41}      & \multicolumn{1}{l|}{19}      & \multicolumn{1}{l|}{43}      & 43      \\ \hline
		style size (px)  & \multicolumn{1}{l|}{$150^2$} & \multicolumn{1}{l|}{$300^2$} & \multicolumn{1}{l|}{$500^2$} & \multicolumn{1}{l|}{$700^2$} & $700^2$ & \multicolumn{1}{l|}{$300^2$} & \multicolumn{1}{l|}{$700^2$} & \multicolumn{1}{l|}{$300^2$} & \multicolumn{1}{l|}{$700^2$} & $500^2$ \\ \hline
		mean rating & \multicolumn{1}{l|}{1.87}    & \multicolumn{1}{l|}{2.02}    & \multicolumn{1}{l|}{2.1}     & \multicolumn{1}{l|}{2.24}    & 2.28    & \multicolumn{1}{l|}{7.85}    & \multicolumn{1}{l|}{7.86}    & \multicolumn{1}{l|}{7.92}    & \multicolumn{1}{l|}{7.98}    & 8.28    \\ \hline
		std rating  & \multicolumn{1}{l|}{1.06}    & \multicolumn{1}{l|}{1.25}    & \multicolumn{1}{l|}{1.28}    & \multicolumn{1}{l|}{0.84}    & 1.25    & \multicolumn{1}{l|}{1.28}    & \multicolumn{1}{l|}{1.38}    & \multicolumn{1}{l|}{1.26}    & \multicolumn{1}{l|}{1.27}    & 1.11    \\ \hline
	\end{tabular}
	\caption{5 styles with the least rating and 5 styles with the greatest rating in the dataset.}
	\label{table:style_ratings}
\end{table*}

Average stylization quality heavily depends on style and its size, as can be seen in Table~\ref{table:style_ratings}. This dependency is much stronger, than from the content image (Table~\ref{table:content_ratings}). 

In particular we emphasize strong impact of style size (irrespective of content and style images), shown in Table~\ref{table:size_ratings}. The size of the style image affects the size of artistic effects on the stylization transferred from the style image. Results in Table~\ref{table:size_ratings} show that small size is detrimental and also quality deteriorates if the style size is too big. We found that $500^2$ pixels is optimal style, providing best results on average. 

Small style size yields small high-frequency artistic  effects on the stylization overwhelming the perception of the original content, as can be seen on Fig.~\ref{fig:st_scale}. On the other hand, if style size is too large, convolutions involved in style transfer methods are not capable to reproduce large artistic patterns, since they have fixed receptive fields and style transfer has negligible effect on the content image, as illustrated on the bottom row of  Fig.~\ref{fig:st_scale_2}.

Style-size combinations with the highest average ratings are presented on Fig.~\ref{fig:style_good_stylizations}. For such cases style image size is not too small (minimal size is $300^2$ pixels) and the image itself contains artistic patterns that are neither too small (to overwhelm content) nor too large (to be missed by convolutions with fixed receptive field). Favorable style images also contain diverse colors and diverse artistic patterns under different angles and of different shapes. Variability in angles allows to accurately preserve edges of the content image thus preserving its recognizability. Large and small shapes of artistic patterns allow favorable stylization of high-frequency content (on foreground) and low-frequency content (on background).

\begin{figure*}[h!t]
	\centering
	\includegraphics[width=1\textwidth]{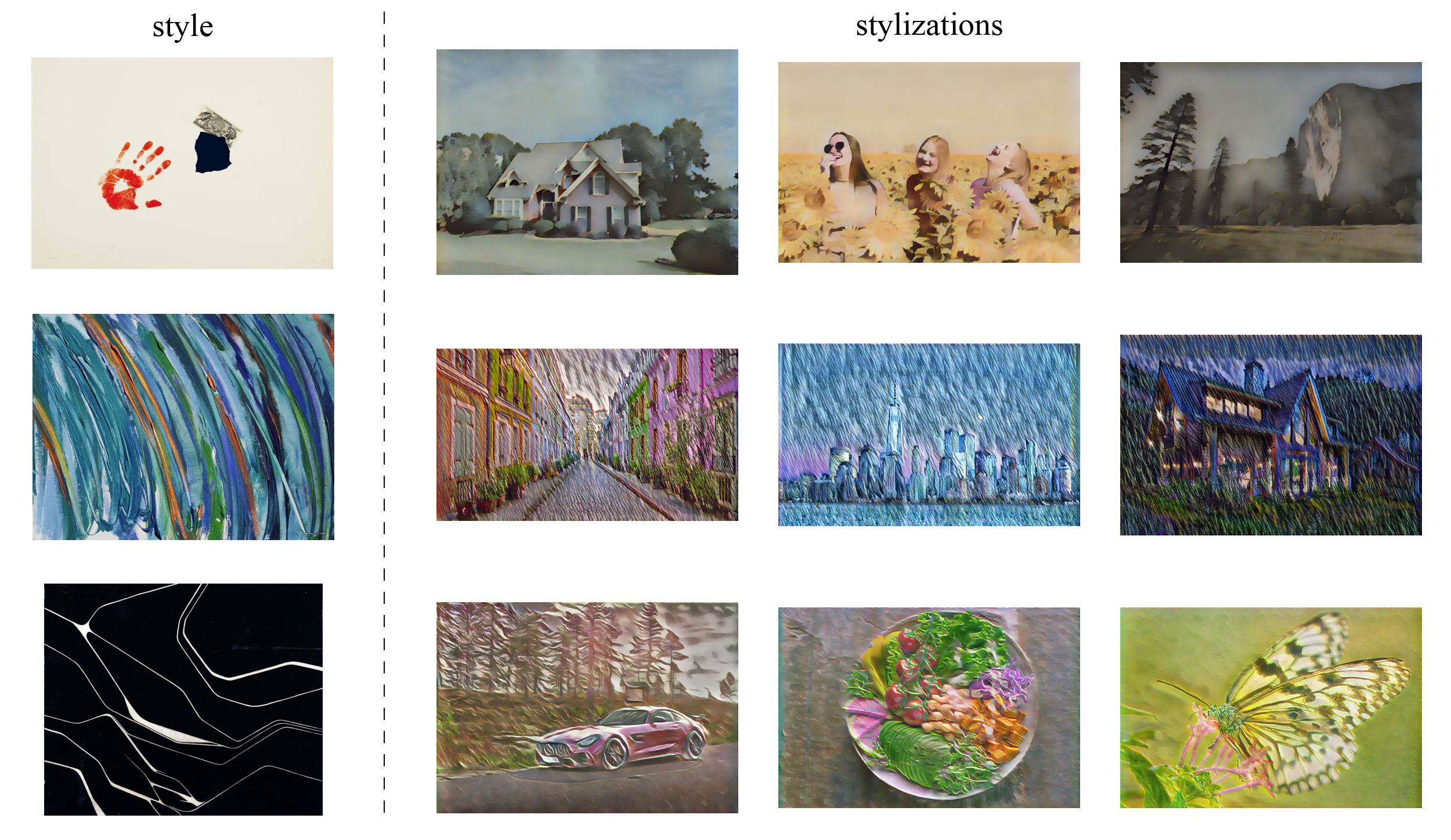}
	\caption{Examples of styles that provide low quality stylizations. Respective sizes of style images are $150^2$, $300^2$ and $500^2$ pixels.}
	\label{fig:style_bad_stylizations}
\end{figure*}

On the other hand, styles with low average ratings, such as shown on Fig.~\ref{fig:style_bad_stylizations}, lack a variety in colors, edges and more complicated textures (the first and the last style). This results in stylizations being over-smoothed and blurs some of the content details. The second unfavorable case, shown in the second row of Fig.~\ref{fig:style_bad_stylizations}, is when style image contains too many high-frequency patterns, especially if they all contain edges at the same angle. Transferred high-frequency patterns overwhelm original content. Additional content information loss occurs, because style edges are at the same angle, wheres content objects contain angles of different angles. Thus most of them are ignored and erased by style transfer, impairing original content recognizability.

\section{Recommendations}  \label{seq:recommendations}

We also highlighted that style transfer can be performed by transferring only artistic patterns without color. Specifically, our dataset was created this way to omit user evaluation bias, favoring particular colors instead of the actual behavior of the style transfer itself.

As the result of a thorough analysis of the dataset, we provide a list of recommendations for generating a high-quality stylization.

\begin{enumerate}
    \item Size of artistic patterns matters. If they are too small, they become unrecognizable and style transfer reduces to photorealism. However, if they are small and vivid, containing pronounced edges, implied artistic effects can suppress the original content information. Also style transfer algorithms have limited capacity to reproduce large artistic patterns, distributed over broad areas. Styles with large smoothly varying patterns over-smooth resulting stylization. 
    \item Effective style should contain a combination of large and small artistic patterns and they should approximately fit to the distribution of high-frequency and low-frequency data on the content image.
    \item Size of the style image itself is important since it controls the average scale of artistic patterns on the image. Thus, style transfer algorithms should be able to work with content and style images of different sizes. 
    \item Style image should contain a variety of textures and diverse colors. Oversimplified styles impose blurring and the loss of important details during stylization.
    \item The style image should contain edges at different angles. In this case style transfer retains the ability to preserve complex contours on the content image, thus maintaining its recognizability.
    \item Particular care should be taken when stylizing close human faces and large smooth areas such as the sky, because style transfer tends to exaggerate even minimal edges and color changes, causing noticeable distortion on the resulting image. 
\end{enumerate}

\section{Conclusion} \label{seq:conclusion}
We presented a new dataset for style transfer, covering various content and style images of different size and containing 10.000 stylizations, rated by three annotators. This dataset can be used to train models, predict optimal content, style and style size combinations. 

Given obtained ratings, we provided analysis how stylization evaluation grades are distributed and which qualitative characteristics and quantitative measures are responsible for forming those grades. We discussed what makes stylization  good or bad and summarized our findings in a concise list of recommendations.

We hope that this work will drive the advancement of style transfer algorithms, making their results more vivid and appealing for the end user.

\section*{Acknowledgments}
This research was performed in the framework of the state task in the field of scientific activity of the Ministry of Science and Higher Education of the Russian Federation, project "Models, methods, and algorithms of artificial intelligence in the problems of economics for the analysis and style transfer of multidimensional datasets, time series forecasting, and recommendation systems design", grant no. FSSW-2023-0004.

\bibliographystyle{unsrt}  
\bibliography{references}  

\end{document}